\pdfoutput=1
\documentclass[11pt]{article}
\usepackage{acl}
\usepackage{times}
\usepackage{latexsym}
\usepackage[T1]{fontenc}
\usepackage[utf8]{inputenc}
\usepackage{microtype}
\usepackage{microtype}
\usepackage{hyperref}
\usepackage{todonotes}
 \usepackage{graphicx}
\usepackage{multirow}
\usepackage{color, colortbl}
\usepackage{enumitem}
\usepackage{rotating}

\title{Response Generation in Longitudinal Dialogues: \\ Which Knowledge Representation Helps?}

\author{Seyed Mahed Mousavi, Simone Caldarella, Giuseppe Riccardi\\
         Signals and Interactive Systems Lab, University of Trento, Italy \\
        \texttt{mahed.mousavi@unitn.it,giuseppe.riccardi@unitn.it}}

\begin{document}
\maketitle
\begin{abstract}
Longitudinal Dialogues (LD) are the most challenging type of conversation for human-machine dialogue systems. LDs include the recollections of events, personal thoughts, and emotions specific to each individual in a sparse sequence of dialogue sessions. Dialogue systems designed for LDs should uniquely interact with the users over multiple sessions and long periods of time (e.g. weeks), and engage them in personal dialogues to elaborate on their feelings, thoughts, and real-life events. In this paper, we study the task of response generation in LDs. We evaluate whether general-purpose Pre-trained Language Models (PLM) are appropriate for this purpose. We fine-tune two PLMs, GePpeTto (GPT-2) and iT5, using a dataset of LDs. We experiment with different representations of the personal knowledge extracted from LDs for grounded response generation, including the graph representation of the mentioned events and participants. We evaluate the performance of the models via automatic metrics and the contribution of the knowledge via the Integrated Gradients technique. We categorize the natural language generation errors via human evaluations of contextualization, appropriateness and engagement of the user.
\end{abstract}

\section{Introduction}


The state-of-the-art dialogue systems are designed for assisting the user to execute a task, holding limited chit-chat conversations with shallow user engagement, or information retrieval over a finite set of topics. The personalization in these systems is limited to a stereotypical user model. This user model is implicitly inferred from conversations with many users, or is limited to a superficial list of persona statements (e.g., "He likes dogs") \cite{zhang2018personalizing}.
The dialogue sessions are disconnected and the shared information across sessions is negligible and close to none.  

Longitudinal Dialogue (LD) is one of the most challenging types of conversation for human-machine dialogue systems. LDs are multi-session interactions that encompass user-specific situations, thoughts, and emotions. Dialogue systems designed for LDs should interact uniquely with each user about personal life events and emotions over multiple sessions and long periods of time (e.g. weeks). Through each session in LDs, the dialogue system must learn about the user's personal space of events and participants and social interactions, and engage the user in personal dialogues regarding their thoughts, feelings, and personal and world events.  

\begin{figure*}[t!]
    \centering
    \includegraphics[width=0.95\textwidth]{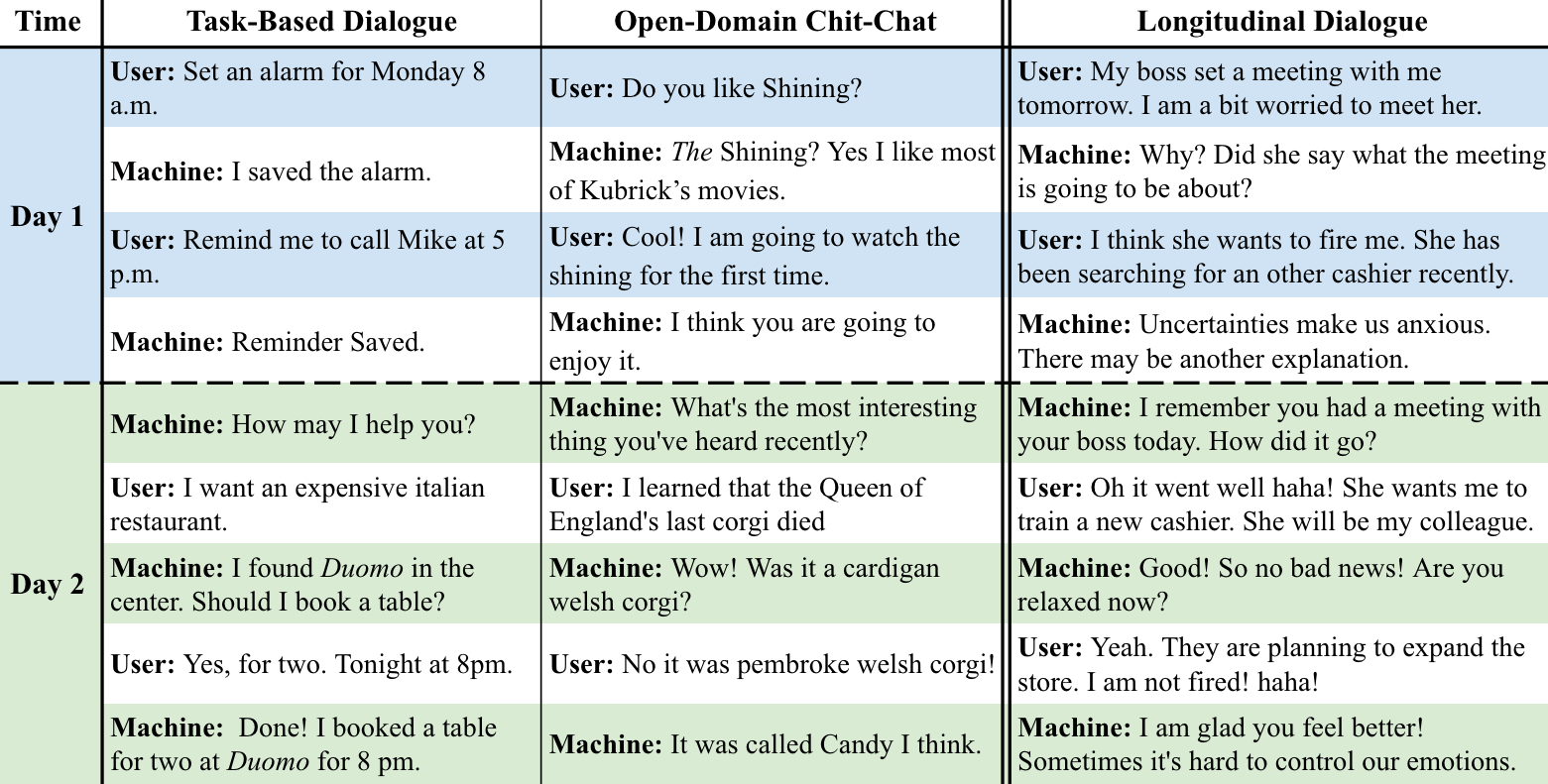}
    \caption{Examples of a task-based dialogue, a chat-chit, and a Longitudinal Dialogue (LD) in two different sessions. The dialogue system for LDs needs to learn about the user in a timely manner and engage her in a personal conversation encompassing her life events, thoughts, and emotions.
    }
    \label{fig:differentdialogues}
\end{figure*}

Figure \ref{fig:differentdialogues} shows an example of three types of human-machine dialogues: task-based, open-domain chit-chat and LD. The user dialogues with the tasked-based dialogue system 
consists of either independent command-and-control exchanges such as on Day 1, or a task-driven dialogue such as on Day 2. The user model in this system is not personal as it adopts a stereotypical model -implicitly - inferred from dialogue corpora with multiple users. In the open-domain chit-chat dialogue, the dialogue does not include the execution of any explicit task, and the model engages the user in a conversation about movies and news. A common characteristic of task-based and open-domain dialogues is the fact that there is no personal information carried to the next dialogue session. The system does not update/modify the user model with each dialogue session and the level of personalization is intact from one interaction to the other (Personalization in the natural language processing and dialogue models could be added based on the voice user interface requirements and could include the exploitation of personal information such as contact directory, preferences, etc.). 

In contrast, the model designed for the LD  must account for three main differences compared to the other two systems; A) the contents of the LD are not about general information or knowledge matters as LDs encompass personal emotions, user and time-specific situations, and participants; B) the sessions are not disconnected dialogues and we can not model them as stand-alone interactions. In contrast, they belong to a multi-session interaction unique to the individual user, where the information shared in each interaction creates a common ground between the machine and the user. For each interaction, the system must engage the user in a dialogue respecting the common ground based on the information shared in the previous interactions, as well as the novel information in the new dialogue history; C) the machine has to extract the personal information presented in the user responses to construct and update the user model and respond coherently. Similar to a natural interaction between human speakers, the model has to gradually become acquainted with the user throughout the dialogues and not from a superficial list of sentence-based persona descriptions.

\begin{figure*}[t!]
    \centering
    \includegraphics[width=0.95\textwidth]{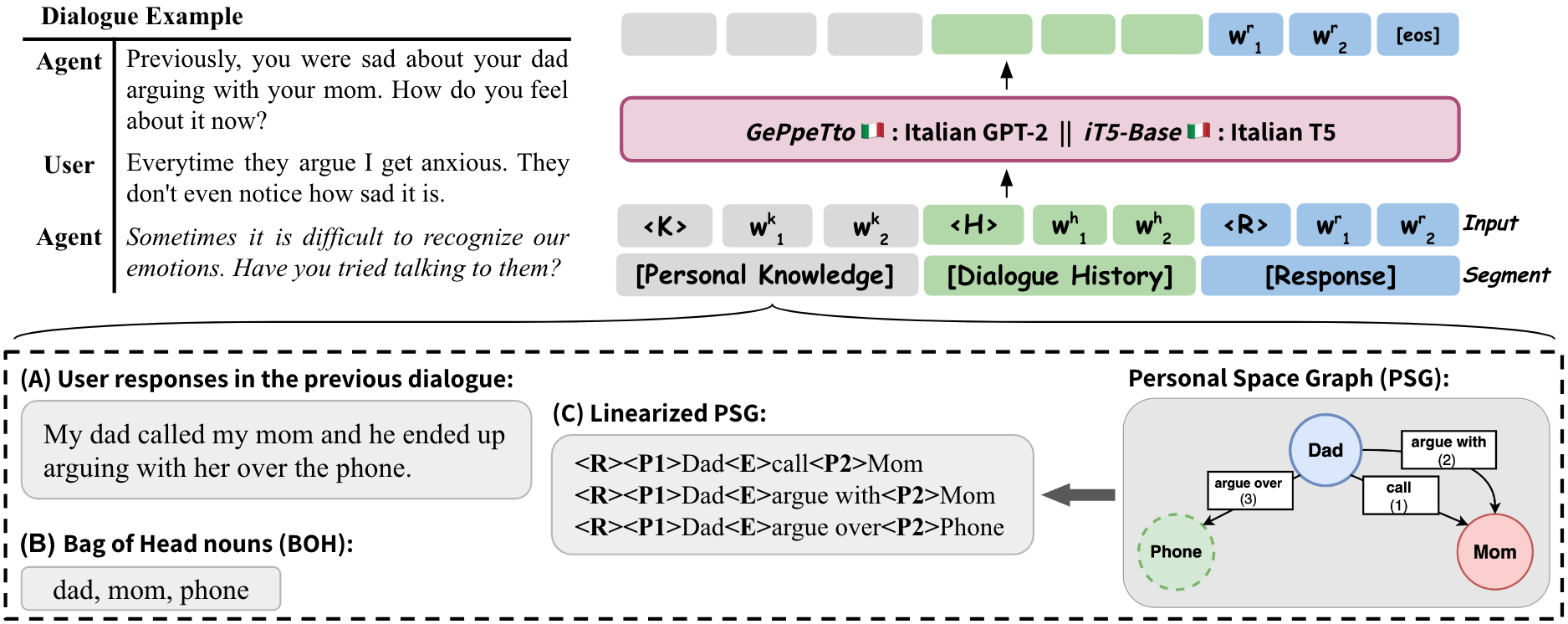}
    \caption{An example of a longitudinal dialogue. The user responses in the previous dialogue session are used as personal knowledge for grounded response generation. The knowledge is presented to the model as A) Unprocessed text (\textit{RAW}); B) Bag of Head nouns (\textit{BOH}); and C) Personal Space Graph (\textit{PSG}) of events and their participants in linearized format. The model then encodes the dialogue history and the knowledge piece and generates a response candidate (the last agent turn in the dialogue example).}
    \label{fig:model}
\end{figure*}

There has been limited research on personal conversations with users over a long period of time. Engaging the user to elaborate on personal situations and emotions is a challenging task and designing appropriate collection/elicitation methodologies is not straightforward. As a result, research on multi-session dialogues resorts to crowd-sourcing datasets with superficial persona statements and pretended longitudinality \cite{goldfish,longtime,updatedbae}. Meanwhile, studies on LDs have been limited to inferring user's attributes such as age and gender \cite{welch2019look}, or next quick-response selection from a candidate set of “yes,” “haha,” “okay,” “oh,” and “nice” \cite{welch2019learning}.

In this work, we study the task of response generation in LDs. Response generation in LDs is subject to appropriateness and accuracy as well as personalization and engagement of the user. The level of personalization in LDs is beyond a set of personal preferences and can not be learned from a limited set of persona statements ("\textit{I like cars}" does not necessarily imply that I like to talk about cars in my interactions). The generated response needs to respect individuals' states, profiles, and experiences that vary among users and dialogue sessions. Therefore, we can not collect a massive knowledge base of user models that can suit all individuals and scenarios. The dialogue system should learn about each user and derive the individual user model through/from the previous dialogue sessions to generate a personal response that is coherent with respect to the dialogue context as well as the previous dialogue sessions. 

We investigate the applicability of general-purpose Pre-trained Language Models (PLM) for grounded response generation in LDs. We study whether PLMs can generate a response that is coherent with respect to the dialogue history and grounded on the personal knowledge the user has shared in previous interactions.  We conversationally fine-tuned two recent PLMs, GePpeTto (GPT-2) \cite{de2020geppetto} and iT5 \cite{sarti2022it5}, using a dataset of LDs about real-life events, feelings, and situations that the user has experienced. We use the responses each individual user shared in the previous dialogue sessions with the system as personal knowledge and evaluate whether grounding the generation on such knowledge results in more appropriate and personal responses. In previously published research on grounded generation, the knowledge sequence is provided to the model as-is. In this work, we experiment with three different representations of the knowledge piece; A) \textit{Raw} as unprocessed text, similar to the previously published research;  B) bag of head nouns as a distilled syntactic representation of the knowledge;  C) graph representation of the events and participants mentioned in the user responses \cite{mousavi2021unsupervised}. An example of a dialogue and different representations of the corresponding personal knowledge is shown in Figure~\ref{fig:model}.

We evaluate the performance of the models and the impact of different knowledge representations through automatic and human evaluations, as well as explainability studies using the Integrated Gradients technique \cite{pmlr-v70-sundararajan17a}. Our contributions can be summarised as follows:
\begin{itemize}[noitemsep,topsep=1pt,parsep=1pt,partopsep=1pt]
    \item To the best of our knowledge this is the first study on the task of response generation in LDs.
    \item We conversationally fine-tune two PLMs with and without grounded response generation on personal knowledge. We study the performance of the models and how different representations of knowledge can affect generation quality.
    \item We evaluate and compare the performance of the models using automatic evaluation, including explainability studies, and human evaluations, including studying the sub-dimensional errors made by each model.
\end{itemize}


\section{Literature Review}

\textbf{Grounded Response Generation} PLMs have achieved comparably well performance for open-domain chit-chats \cite{zhang-etal-2020-dialogpt}, goal-oriented agents \cite{thulke2021adapting} and question answering \cite{zhao-etal-2020-knowledge-grounded}. However, such models can generate inappropriate and/or generic responses which can lead to ethical problems and low user engagement \cite{zhang-etal-2020-dialogpt}. Research to address this problem and improve the generation quality includes grounding the generation on external knowledge content. The selection of the knowledge source to ground the generation has been studied as an individual component \cite{hedayatnia-etal-2020-policy}, as well as a joint task along with response generation \cite{zhao-etal-2020-knowledge-grounded,huang-etal-2021-plato}. 

\textbf{Personal Dialogue} Research on personalized response generation has focused on persona descriptions and synthetic sets of user preferences and profiles.
\citet{zhang2018personalizing} collected Persona-Chat dataset of open-domain dialogues using crowd workers, where the workers were instructed to impersonate as speakers with synthetic personas of 5 sentences. This dataset has been studied for personal response generation by fine-tuning PLMs \cite{wolf2019transfertransfo,kasahara-etal-2022-building}, by learning the users' persona from the dialogues samples rather than the persona descriptions \cite{madotto-etal-2019-personalizing}, or investigating different representations of persona statements \cite{huang2022personalized}. While the mentioned work focused on personalization in open-domain dialogues, \citet{joshi2017personalization} generated profiles consisting of gender, age, and food preference permutations for the user side in restaurant booking dialogues, which was used in another work \cite{siddique2022personalizing} to generate personalized responses in a task-based dialogue.

\textbf{Multi-session Dialogue} Studies on multi-session dialogues have been limited to simulated longitudinality and superficial persona. \citet{goldfish} extended the Persona-Chat dataset to a multi-session chat dataset with 4 to 5 sessions, by instructing crowd-workers to impersonate the role of returning dialogue partners in the first session (extracted from the Persona-Chat dataset) after a random amount of time. The workers were explicitly asked not to discuss any personal and real-life matters but play the role defined by the persona statements. This approach was further used by \citet{updatedbae} to extend an existing dataset of persona chats in Korean to multi-session dialogues. \citet{longtime} proposed a framework for persona memory in multi-session dialogues and collected a dataset of persona chats in Chinese via crowd workers. 


\section{Experiments}



\subsection{Dataset}

The dataset of LDs used in this work \cite{mousavi-etal-2021-like} consists of two dialogue sessions for each individual user. The first dialogue session is a set of personal human-machine conversations with real users encompassing their personal life events and emotions. These dialogues are collected from a group of 20 Italian native speakers receiving therapy to handle their distress more effectively. Throughout the interaction, the machine prompts the user to engage her in the recollection of daily life events the user has experienced, while the user shares details about the events and participants that have activated her emotions by answering a set of questions.

For each user, the first session is then followed by a follow-up dialogue. These dialogues were elicited from 4 psychotherapists and 4 trained annotators supervised by the psychotherapists. In the second dialogue session, the user tends to share more details about her feelings and the possible evolution of the previously mentioned events. Meanwhile, the listener provides personal suggestions and asks questions to expand or disambiguate previously stated facts or feelings. A mock-up example of a second dialogue session and the corresponding user response in the previous dialogue is shown in Figure~\ref{fig:model}. This dataset consists of 800 2-session LDs in the mental health domain with an average of 5 turns per dialogue. 

\subsection{Models}
 
We fine-tuned two state-of-the-art PLMs using the dataset of LDs. 

\textbf{GePpeTto: Italian GPT-2} The first model we experimented with is GePpeTto \cite{de2020geppetto}, a PLM based on GPT-2 small (12 layers of decoder, 117M parameters) \cite{radford2019language}, trained for the Italian language (13 GB corpus size). We fine-tuned the model using AdamW optimizer \cite{loshchilov2017decoupled} with an early-stopping wait counter equal to 3 and a history window of 2 last turns.

\textbf{iT5: Italian T5} The second PLM in our experiments is iT5 \cite{sarti2022it5}, a PLM based on T5 \cite{raffel2020exploring}, trained on the Italian portion of mC4 corpus (275 GB corpus size). We experimented with iT5-Small (12 layers, 60M parameters) and iT5-Base (24 layers, 220M parameters) \footnote{We were unable to use iT5-Large due to lack of GPU memory}. We fine-tuned this model class using AdaFactor optimizer \cite{vaswani2017attention} with early stopping wait counter equal to 3 and a history window of 4 last turns.

\subsection{Grounded Response Generation}
For each user, we extracted her responses in the first dialogue session as personal knowledge to ground the response generation for the second dialogue session. We experimented with three representations of the knowledge piece:

\begin{itemize}[noitemsep,topsep=2pt,parsep=2pt,partopsep=2pt]
    \item \textbf{(A) RAW}: We provide the responses of the user in the previous dialogue as an unprocessed knowledge piece. The average length of knowledge with this representation is 126.7 tokens.
    
    \item \textbf{(B) Bag of Head nouns (BOH)}: We automatically parse the user responses \footnote{the dependency parser used is spaCy: \href{spacy.io}{spacy.io} } and extract the head nouns as a distilled syntactic representation of the knowledge. 

    \item \textbf{(C) Personal Space Graph (PSG)}: We represent the knowledge by the personal graph of the events and participants mentioned by the user \citet{mousavi2021unsupervised}. The predicates in a sentence represent an event, and its corresponding noun dependencies (subject, object) represent the participants. In this graph, the participants are the nodes while the predicates are the relations (edges) among the participants. We obtain a linear representation of the graph using an approach inspired by \citet{ribeiro-etal-2021-investigating} in which the authors observed that providing a linearized representation of the graph to the PLMs results in outperforming the models with a graph-specific structural bias for the task of graph-to-text generation. 
\end{itemize}

\section{Evaluations}
The fine-tuning of the models was done using 80\% of the dialogues (640 second-session dialogues, 1284 samples with different turn levels), while the remaining data was split into 10\% (80 dialogues, 160 samples with different turn levels) as the validation set for parameter engineering and early-stopping, and 10\% as unseen test set. Each split was sampled at the dialogue level to guarantee no history overlap among splits. An example of a second dialogue session and the generated responses are presented in Appendix Table \ref{tab:dialogueex}.

\subsection{Automatic Evaluation}
The results of the automatic evaluation of the models is presented in Table~\ref{table:autoeval}. 
The perplexity scores cannot be used to compare the performance between GePpeTto and iT-5 model classes as the vocabulary distributions in the pre-training phase of the two PLMs are not identical. However, the scores are comparable among iT5 variations as the same model class pre-trained using the same data. In fact, the perplexity scores indicate that iT5-Base demonstrates a better performance than iT5-Small in all combinations with knowledge representations. Therefore, we select iT5-Base among the iT5 models and focus the rest of the analysis on GePpeTto and iT5-Base.

\begin{table}[t]
\centering

\begin{tabular}{ccc}
{\textbf{Models}} &\textit{\textbf{nll}} & \textit{\textbf{ppl}}\\
\hline \hline
\multicolumn{1}{l}{\textit{GePpeTto}} &2.76&15.84\\
\hspace{0.25cm}$_{+RAW Knowl.}$ &2.79&16.33\\
\hspace{0.25cm}$_{+BOH Knowl.}$ &2.85&17.38\\
\hspace{0.25cm}$_{+PSG Knowl.}$ &2.77&16.06\\
\hline \hline
\multicolumn{1}{l}{\textit{iT5-Small}} &2.18&8.84\\
\hspace{0.25cm}$_{+RAW Knowl.}$ &2.19&8.95\\
\hspace{0.25cm}$_{+BOH Knowl.}$  &2.18&8.88\\
\hspace{0.25cm}$_{+PSG Knowl.}$  &2.19&8.93\\
\hline
\multicolumn{1}{l}{\textit{iT5-Base}} &2.05&7.79\\
\hspace{0.25cm}$_{+RAW Knowl.}$      &2.04&7.70\\
\hspace{0.25cm}$_{+BOH Knowl.}$     &2.12&8.40\\
\hspace{0.25cm}$_{+PSG Knowl.}$    &2.09&8.07\\
\hline \hline
\end{tabular}
\caption{Automatic evaluation of the models indicates that incorporating the knowledge slightly increases the models' perplexity (Perplexity scores can not be compared among models since the vocabulary distributions of pre-training data are not identical).}
\label{table:autoeval}
\end{table}

\begin{figure}[t!]
    \centering
    \includegraphics[scale=0.036]{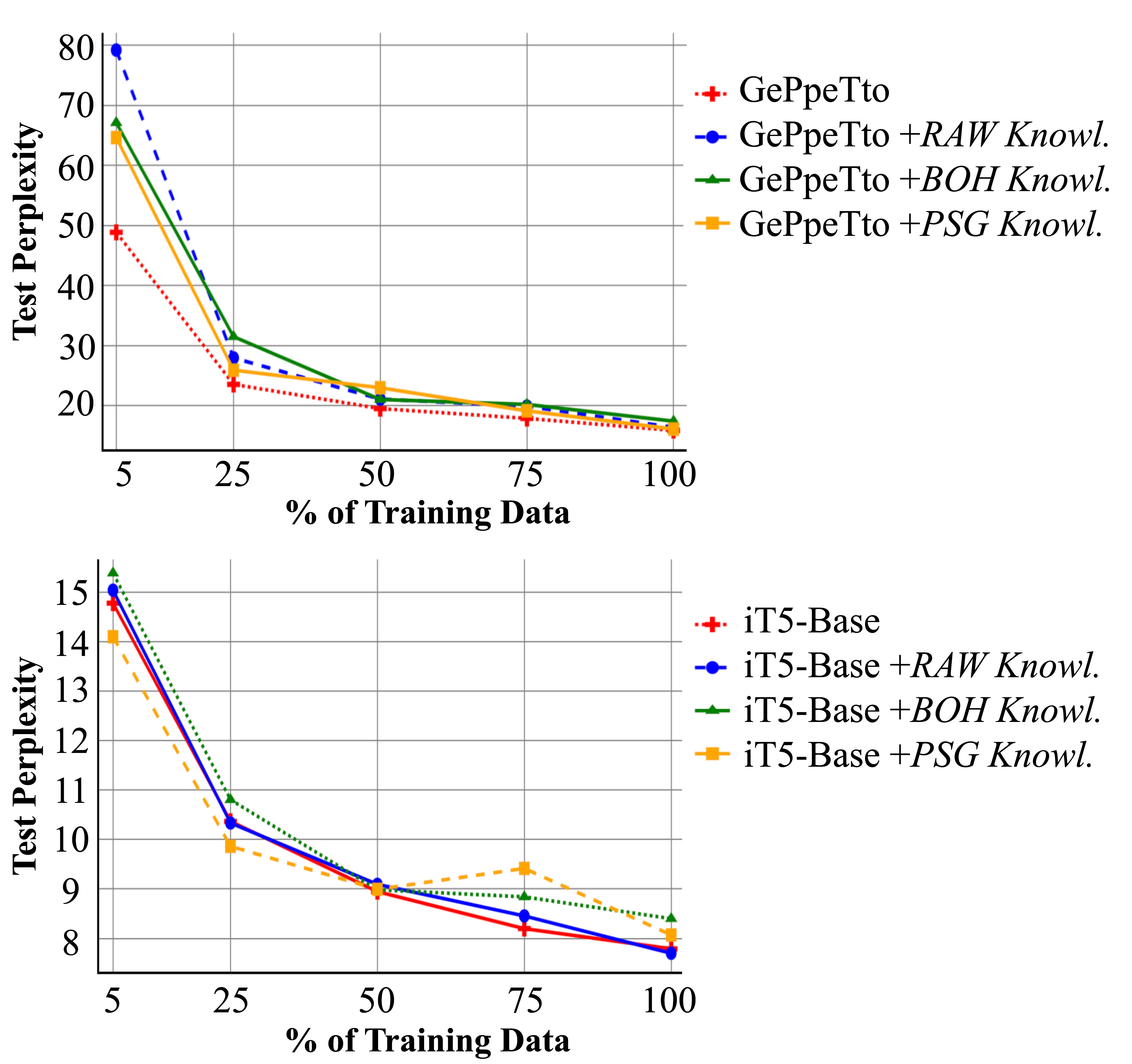}
    \caption{Perplexity score trends of the models over increasing size of the training set. The performance of GePpeTto variations is considerably improved after observing 50\% of the fine-tuning training set.}
    \label{fig:trend}
\end{figure}

Considering the small size of the LD dataset compared to the data used in the pre-training phase, we studied the impact of fine-tuning the models by optimizing the models over increasing size of the training set. The extension of the training set was gradual (the small portions are subsets of the big portions) and the performance of models was evaluated by measuring the perplexity score on the unseen test set. The results are presented in Figure~\ref{fig:trend}.  The performance of both models is improved considerably after observing the first 25\% and 50\% of the train set, thus the fine-tuning has been more effective. However, in the second half of the data, both models show a steady trend while iT5-Base achieves a gradual improvement.

\begin{figure}
    \centering
    \includegraphics[width=0.45\textwidth]{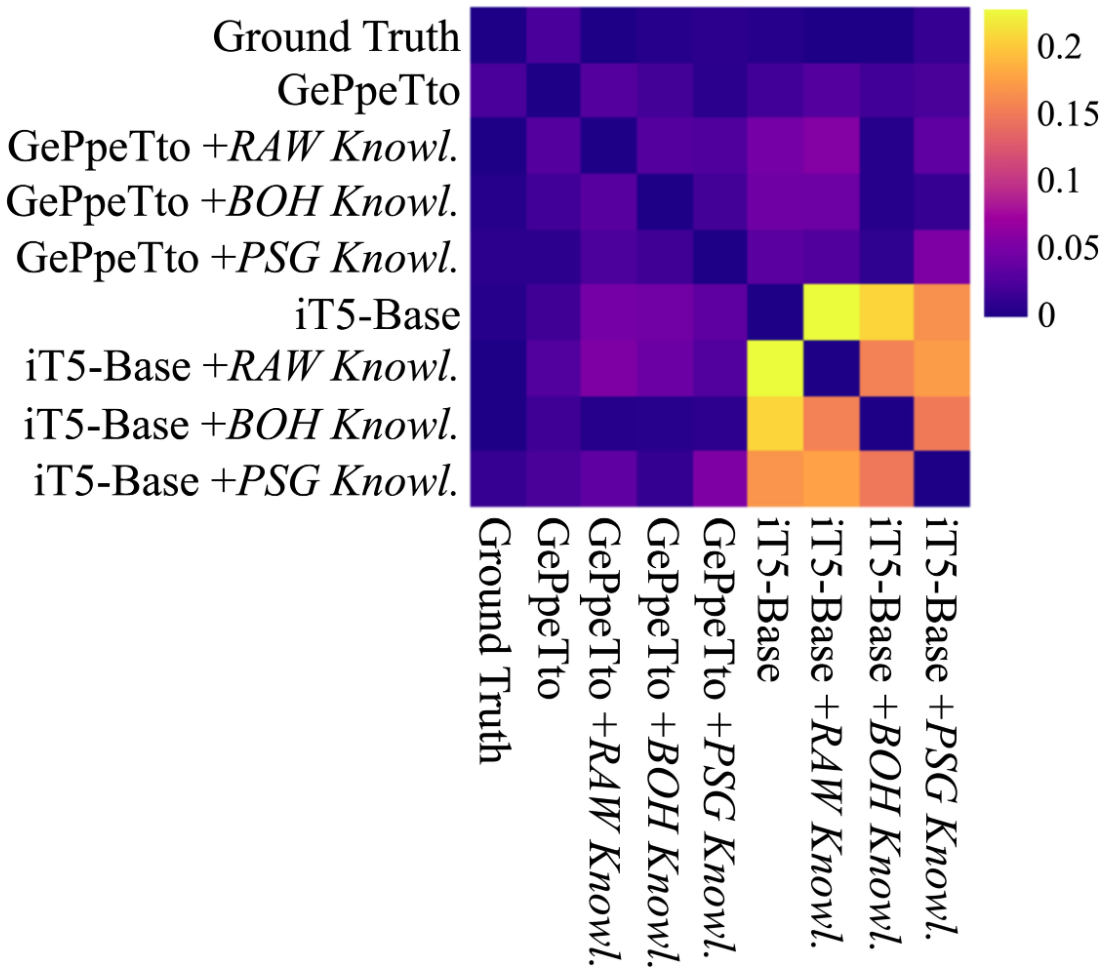}
    \caption{Lexical similarity among generated responses measured by BLEU-4 score. The results indicate a higher similarity among the responses generated by iT5-Base models.}
    \label{fig:blue}
\end{figure}

To investigate the impact of grounding on the response lexicalization of the models, we measured the diversity in the generated responses for the test set samples via BLEU-4 score, Figure~\ref{fig:blue}. We observed that there is a higher similarity among responses generated by iT5 models, while the responses generated by GePpeTto variations are more diverse. A similar finding has been observed in the literature about the performance of auto-regressive models compared to encoder-decoder architectures regarding novelty in sequence generation \cite{tekiroglu-etal-2022-using, bonaldi-etal-2022-human}. Further, responses generated by iT5-Base with \textit{BOH} and \textit{PSG} representations have the lowest lexical similarity. The responses with the highest lexical similarity are generated by iT5-Base with no grounding and \textit{RAW} representation. Nevertheless, there is a negligible lexical similarity between the generated responses and the ground truth.

\begin{table*}
\centering
\small
\resizebox{\textwidth}{!}{
\begin{tabular}{ccc|cccc}

\multirow{3}{*}{\textbf{Models}} & \multicolumn{2}{c}{\textbf{}} & \multicolumn{4}{c}{\textbf{Human Evaluation}} \\
\cline{2-7}
& \multirow{2}{*}{\textit{\textbf{nll}}} & \multirow{2}{*}{\textit{\textbf{ppl}}} & \multirow{2}{*}{\textbf{Correctness}} & \multirow{2}{*}{\textbf{Appropriateness}} & \multirow{2}{*}{\textbf{Contextualization}} & \multirow{2}{*}{\textbf{Listening}} \\
&&&&&&\\
\hline
\hline
\multicolumn{1}{l}{\textit{Ground Truth}} &-&-& 97.62\%&100.0\%&97.62\%&97.62\%\\
\hline
\multicolumn{1}{l}{\textit{GePpeTto}} &2.76&15.84&
83.33\%&\textbf{66.67\%}&\textbf{69.05\%}&\textbf{64.29\%}\\
\hspace{0.25cm}$_{+RAW Knowl.}$ &2.79&16.33&83.33\%&59.52\%&57.14\%&57.14\%\\
\hspace{0.25cm}$_{+BOH Knowl.}$ &2.85&17.38&\textbf{92.86\%}&45.24\%&52.38\%&42.86\%\\
\hspace{0.25cm}$_{+PSG Knowl.}$ &2.77&16.06&90.48\%&54.76\%&64.29\%&50.00\%\\
\hline
\multicolumn{1}{l}{\textit{iT5-Base}} &2.05&7.79& \textbf{100.0\%}&66.67\%&73.81\%&66.67\%\\
\hspace{0.25cm}$_{+RAW Knowl.}$      &2.04&7.70& 85.71\%&{80.95\%}&80.95\%&76.19\%\\
\hspace{0.25cm}$_{+BOH Knowl.}$     &2.12&8.40&92.86\%&\textbf{80.95\%}&85.71\%&{83.33\%}\\
\hspace{0.25cm}$_{+PSG Knowl.}$    &2.09&8.07& 95.24\%&73.81\%&\textbf{90.48\%}&\textbf{83.33\%}\\
\hline \hline
\end{tabular}}
\caption{Human Evaluation of the fine-tuned models. The results show the impact of different representations of the knowledge source for grounded response generation in LDs. Refined representations of the knowledge (\textit{BOH} and \textit{PSG}) generally result in better performances than \textit{RAW} representation.}
\label{table:he}
\end{table*}

\subsection{Human Evaluation}

We sampled 50\% of the unseen test set (42 dialogue histories, 80 samples with different turn levels) and evaluated the generated responses via human judges. We evaluated the responses according to four criteria using the protocol proposed by \citet{mousavi-etal-2022-eval}:
\begin{itemize}[noitemsep,topsep=2pt,parsep=2pt,partopsep=2pt]
    \item \textbf{\textit{Correctness}}: evaluating grammatical and syntactical structure of the response.
    \item \textbf{\textit{Appropriateness}}: evaluating the response to be a proper and coherent continuation with respect to the dialogue history. 
    \item \textbf{\textit{Contextualization}}: evaluating whether the response refers to the context of the dialogue (not generic) or it consists of non-existing/contradicting information (hallucination cases).
    \item \textbf{\textit{Listening}}: whether the generated response shows that the speaker is following the dialogue with attention.
\end{itemize}

The annotators were asked to evaluate the response candidates and select a decision for each criterion from a 3-point Likert scale as positive (eg. Correct, Appropriate), negative (eg. Not Correct, Not Appropriate), and "I don’t know".
We recruited 35 native Italian crowd-workers through Prolific crowd-sourcing platform\footnote{ Prolific: \url{https://www.prolific.co/}}. The workers were asked to perform a qualification task consisting of evaluating 5 samples (sampled from the validation set) in an identical setting to the main task. For the main evaluation, each crowd-worker annotated 3 response candidates for 10 dialogue histories, and each sample was annotated by 7 crowd-workers. We also asked the annotators to motivate their decisions for appropriateness and contextualization criteria by providing an explanation to point out possible errors in the generated response. Moreover, the ground truth was also included in the candidate set to be evaluated.

The Inter Annotator Agreement (IAA) level measured by Fleiss’ $\kappa$, presented in Appendix Table~\ref{table:agreements}, indicates high levels of subjectivity and complexity in \textit{Contextualization} criterion, suggesting that it has been difficult for the annotators to assess this aspect of the responses. 

The results of the human evaluation of responses are presented in Table~\ref{table:he} (the scores are obtained by majority voting). The evaluation of GePpeTto models shows that grounding generally worsens the performance of GePpeTto, regardless of the representation format, as the best performance is achieved by GePpeTto with no knowledge grounding. Nevertheless, \textit{BOH} and \textit{PSG} representations slightly improve the grammatical correctness of this model. The highest level of \textit{Contextualization} among grounded GePpeTto models is achieved by \textit{PSG} representation. Regarding iT5-Base variations, the results indicate that grounding improves the models' performance considerably with respect to \textit{Appropriateness}, \textit{Contextualization}, and \textit{Listening}. However, it decreases the model's \textit{Correctness} with the highest decrease caused by \textit{RAW} representation. \textit{PSG} representation achieves the highest level of \textit{Contextualization} and \textit{Listening} overall, besides the highest level of \textit{Correctness} among grounded models. Therefore, refined representations of the knowledge (\textit{BOH} and \textit{PSG}) generally result in better performances compared to \textit{RAW} representation. Nevertheless, there is still a huge gap between the performance of the best-performing model and the ground truth, suggesting the grounded PLMs are not suitable dialogue models for LDs in the mental health domain.

\begin{figure*}[t!]
    \centering
    \includegraphics[width=0.95\textwidth]{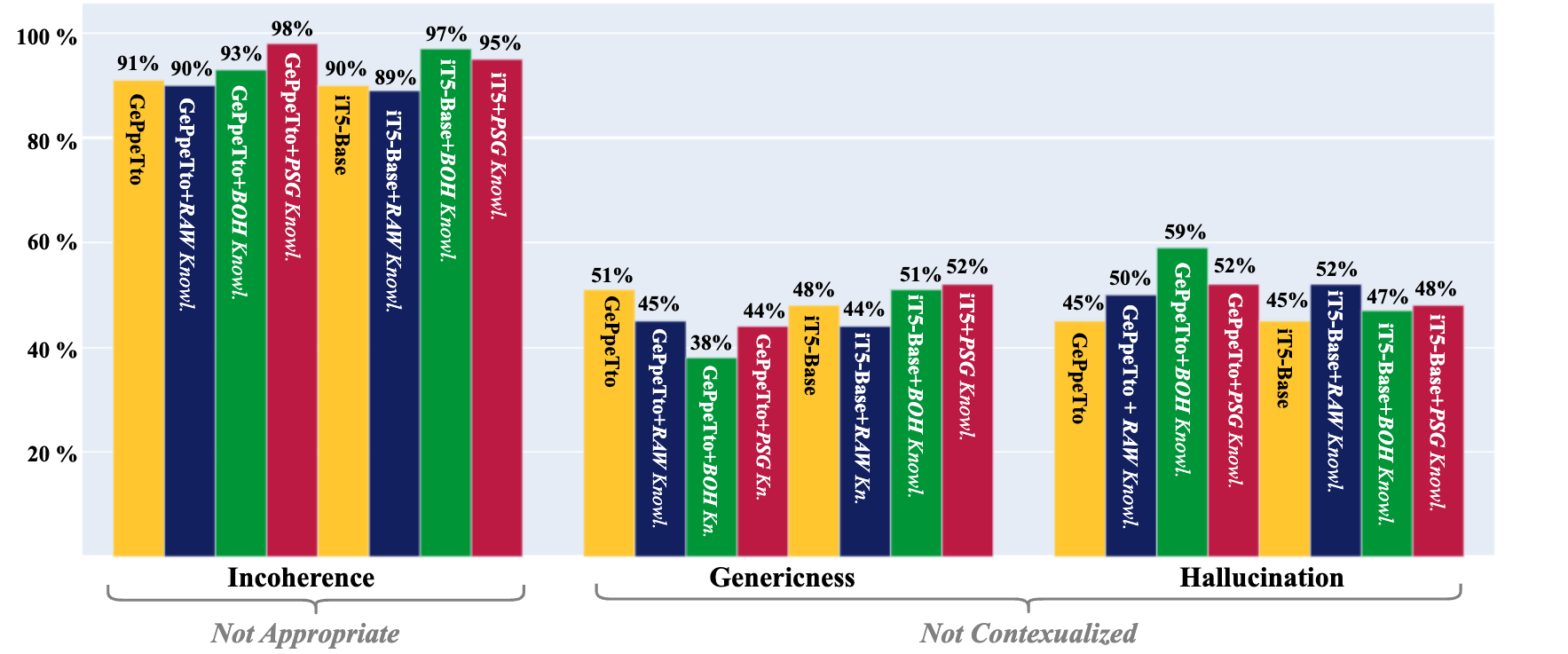}
    \caption{Explanations provided by the crowd-workers to motivate their negative judgments in \textit{Appropriateness} and \textit{Contextualization} criteria, represented by the percentage of the times the error category (x-axis) was selected. The figure is obtained by considering all the votes (i.e. not majority voting). Note that the labels are not mutually exclusive.}
    \label{fig:explanations}
\end{figure*}

To gain better insight into the errors made by each model, we investigated the reasons provided by the annotators for their judgments. These results, presented in Figure \ref{fig:explanations}, are complementary to the evaluation decisions, Table \ref{table:he}, and point out the errors that resulted in the negative evaluation of a response by the annotators. The analysis shows that grounding reduces the cases of genericness in rejected responses by GePpeTto while it slightly escalates this issue in iT5-Base rejected responses. Moreover, the rejected responses of iT5-Base with \textit{RAW} representation were more hallucinated than other representations. Nevertheless, grounding does have any positive impact on the cases of incoherence in rejected responses of the PLMs.

\subsection{Generation Explainability}

According to the human evaluation results, iT5-Base with knowledge grounding achieves the best performance among PLMs. We investigated the contribution of personal knowledge and different representations on the performance of the model at inference time. We studied the attribution scores of the input tokens using the Integrated Gradients technique \cite{pmlr-v70-sundararajan17a,inseq} based on backward gradient analysis. We experimented with two thresholds for the attribution scores: 
\begin{itemize}[noitemsep,topsep=2pt,parsep=2pt,partopsep=2pt]
    \item \textbf{Positive Contribution}: Based on the assumption that elements with positive scores have a positive influence on the model's performance, we investigated the tokens with positive attribution scores, However, tokens with small attribution scores have negligible contributions and thus this analysis can be noisy.
    \item \textbf{Significant Contribution}: To identify the tokens with significant contributions to the generation, we selected the top-25\% of the tokens in the input sequence (knowledge and history) according to their attribution score. We then investigated what portion of these tokens belong to each segment of the input vector. For a fair comparison, the values are normalized over the segment length.
\end{itemize}

According to Positive Contribution analysis, 74\% of the tokens in the \textit{RAW} representation have a positive contribution to the generation with the majority (30\%) of tokens being verbs and nouns. This percentage for \textit{BOH} (Bag of Head Nouns) representation changes to 79.0\%. This result suggests the importance of nouns for the model inference. Regarding the \textit{PSG} representation, 55.6\% of the tokens have a positive contribution to the generation (excluding the tags used for linearization), with the majority (68\%) of tokens being events rather than participants.

The analysis of the tokens with significant contributions is presented in Table~\ref{table:ig}. Regarding the model with \textit{RAW} representation, the percentage of tokens with high attribution scores is almost balanced between the knowledge and history segments. However, for the models with refined representations of knowledge (\textit{BOH} and \textit{PSG}), the dialogue history contains moderately more significantly contributing tokens.

\begin{table}[!t]
\centering

\begin{tabular}{ccc}
{\textbf{Models}}  &\textit{\textbf{Knowl.}} & \textit{\textbf{History}}\\
 
\hline \hline
\multicolumn{1}{l}{\textit{iT5-Base}} &&\\
\hspace{0.1cm}$_{+RAW Knowl.}$   & 44.6\%&55.4\%\\
\hspace{0.1cm}$_{+BOH Knowl.}$   & 39.5\%&60.5\% \\
\hspace{0.1cm}$_{+PSG Knowl.}$   & 38.7\%&61.3\%\\
\hline \hline
\end{tabular}
\caption{Percentage of tokens with significant contribution to the generation (top-25\%) in knowledge and history segments of the input vector for each model.}
\label{table:ig}
\end{table}

\section{Conclusion}

We studied the task of response generation in Longitudinal Dialogues (LD), where the model should learn about the user's thoughts and emotions from the previous dialogue sessions and generate a personal response that is coherent with respect to the user profile and state, the dialogue context, as well as the previous dialogue sessions. We fine-tuned two state-of-the-art PLMs for Italian, using a dataset of LDs in the mental health domain. We experimented with grounded generation using user responses in the previous dialogue session as user-specific knowledge. We investigated the impact of different representations of the knowledge, including a graph representation of personal life events and participants mentioned previously by the user. 

Our evaluations showed there is still a huge gap between the performance of the general-purpose PLMs with knowledge grounding and the ground truth. Nevertheless, we observed that a) refined representations of the knowledge (such as \textit{BOH} and \textit{PSG}) can be more informative and less noisy for a grounded generation; b) the encoder-decoder model exhibited more diversity in the outputs compared to the auto-regressive model; c) knowledge grounding reduces the cases of genericness in response, though it can result in more hallucinated responses. 

\section*{Limitations}

The dataset used in this work is in Italian and there may be language-specific limitations in the model performance. GePpeTto is the only candidate for auto-regressive models for the Italian language at the time of this research. Therefore, its performance may be limited due to the small number of parameters. We were unable to experiment with iT5-Large model due to computation power limitations.

\bibliography{custom}
\bibliographystyle{acl_natbib}

\onecolumn
\appendix

\section*{Appendix}
\label{sec:appendix}

\begin{table*}[h!]
\centering
\begin{tabular}{ccccc|c}
\multirow{2}{*}{\textbf{Models}} & \multicolumn{5}{c}{\textbf{Inter Annotator Agreement Level measured by Fleiss'$\kappa$}} \\
 \cline{2-6}&{\textbf{\textit{Appropriateness}}} & {\textbf{\textit{Contextualization}}} &  {\textbf{\textit{Correctness}}} & {\textbf{\textit{Listening}}} & \textbf{IAA per Model} \\ \hline \hline
\multicolumn{1}{l}{\textit{GePpeTto}} &0.27 &0.14&0.64&0.15& \multirow{1}{*}{$0.32\footnotesize{\pm0.10}$}\\
\hspace{0.25cm}$_{+RAW Knowl.}$  &0.42&0.22 &0.36&0.27&\multirow{1}{*}{$0.36\footnotesize{\pm0.11}$}\\
\hspace{0.25cm}$_{+BOH Knowl.}$  &0.23&0.05 &0.31&0.11&\multirow{1}{*}{$0.27\footnotesize{\pm0.05}$}\\
\hspace{0.25cm}$_{+PSG Knowl.}$  &0.30&0.39 &0.34&0.26&\multirow{1}{*}{$0.42\footnotesize{\pm0.06}$}\\
\hline
\multicolumn{1}{l}{\textit{iT5-Base}} &0.24&0.19&0.06&0.18&\multirow{1}{*}{$0.27\footnotesize{\pm0.04}$}\\
\hspace{0.25cm}$_{+RAW Knowl.}$&0.18 &0.03&0.30&0.21 &\multirow{1}{*}{$0.19\footnotesize{\pm0.06}$}\\ 
\hspace{0.25cm}$_{+BOH Knowl.}$&0.21 &0.17&0.58&0.24 &\multirow{1}{*}{$0.26\footnotesize{\pm0.09}$}\\ 
\hspace{0.25cm}$_{+PSG Knowl.}$&0.17 &0.06&0.27&0.14 &\multirow{1}{*}{$0.19\footnotesize{\pm0.12}$}\\ \hline \hline
\textbf{IAA per} &\multirow{1}{*}{$0.31\footnotesize{\pm0.09}$}
                 &\multirow{1}{*}{$0.20\footnotesize{\pm0.06}$}
                 &\multirow{1}{*}{$0.43\footnotesize{\pm0.20}$}
                 &\multirow{1}{*}{$0.25\footnotesize{\pm0.10}$}
                 &\multirow{2}{*}{-} \\
\textbf{Dimension} &  \textcolor{cyan}{\textbf{Fair}}& \textcolor{red!80}{\textbf{Poor}}&\textcolor{teal!80}{\textbf{Moderate}}&\textcolor{cyan}{\textbf{Fair}}& \\
 \hline
\end{tabular} 
\caption{Inter-Annotator Agreement (IAA) level calculated by Fleiss' $\kappa$ for each model and criterion. Low IAA level for \textit{Contextualization} suggests a high level of subjectivity in this criterion.}
\label{table:agreements}
\end{table*}

\begin{table*}[h!]
    \centering
    \includegraphics[scale=0.65]{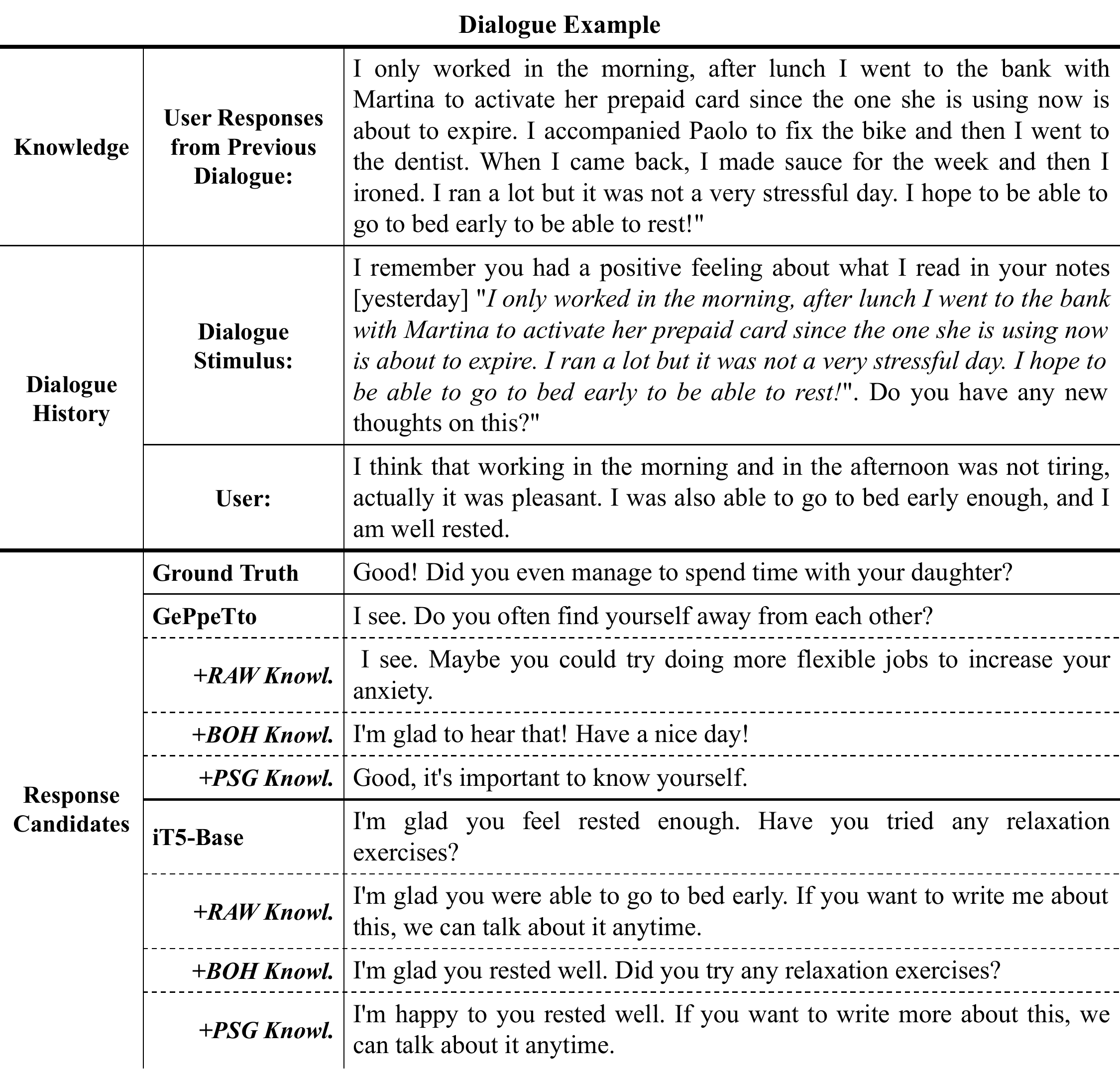} 
    \caption{Example of a second dialogue session, the corresponding user turns in the first session as personal knowledge, and the generated responses (English translation).}
    \label{tab:dialogueex}
\end{table*}

\end{document}